\documentclass[twocolumn, switch]{article}
\usepackage{preprint}
\usepackage{graphicx}
\usepackage{caption}
\usepackage{algorithm}
\usepackage{algorithmic}
\usepackage{amsmath} 
\usepackage{amsfonts}
\usepackage{bbm}
\usepackage{dsfont}

\usepackage[style=authoryear-ibid,backend=biber]{biblatex}
\usepackage{authblk}

\author[1]{Giovanni Charles}
\author[1]{Timothy M. Wolock}
\author[1]{Peter Winskill}
\author[1]{Azra Ghani}
\author[2*]{Samir Bhatt}
\author[3]{Seth Flaxman}

\affil[1]{Imperial College London}
\affil[2]{University of Copenhagen}
\affil[3]{Department of Computer Science, University of Oxford}
\affil[*]{Corresponding Author - s.bhatt@ic.ac.uk}

\title{Seq2Seq Surrogates of Epidemic Models to Facilitate Bayesian Inference}

\begin{document}

\twocolumn[ 
  \begin{@twocolumnfalse} 
  
\maketitle

\begin{abstract}
Epidemic models are powerful tools in understanding infectious disease. However, as they increase in size and complexity, they can quickly become computationally intractable. Recent progress in modelling methodology has shown that surrogate models can be used to emulate complex epidemic models with a high-dimensional parameter space. We show that deep sequence-to-sequence (seq2seq) models can serve as accurate surrogates for complex epidemic models with sequence based model parameters, effectively replicating seasonal and long-term transmission dynamics. Once trained, our surrogate can predict scenarios a several thousand times faster than the original model, making them ideal for policy exploration. We demonstrate that replacing a traditional epidemic model with a learned simulator facilitates robust Bayesian inference.
\end{abstract}

\vspace{0.35cm}

  \end{@twocolumnfalse} 
] 

\section{Introduction}

Epidemic models are an essential tool in the study of infectious disease dynamics \cite{eaton_estimation_2019, world_health_organization_global_2021,djaafara_quantitative_2021}. By modelling new infections as a function of current infections, they leverage the fundamental dynamics to predict epidemic trajectories accurately with relatively few parameters. Recently, they have risen to unprecedented public prominence through their use during the SARS-CoV-2 (coronavirus) pandemic \cite{birrell_real-time_2021}.

In most cases, we do not observe these dynamics or model parameters directly, so we must infer them from what data we do measure (e.g. deaths, positive tests, etc.). In order to produce a set of predictions, we need to infer the unknown parameters in the model, a process that quickly becomes infeasible at the scale needed for methods like Hamiltonian Monte Carlo (HMC) \cite{neal_mcmc_2011} for even moderately complex models. Further, because the current state of the model is a function of all past states, complex correlation structures can arise between time-varying parameters, making sampling difficult.

\cite{griffin_reducing_2010} proposed an epidemiologically motivated Individual-Based Model (IBM) of malaria transmission. It simulates infection at an individual-level based on exposure to three species of infectious mosquito, immunity and a variety of interventions. This model is invaluable for projecting the effects of policy on malaria prevalence , for example, \cite{winskill_us_2017}, projected the effects of a 44\% funding cut from a major global donor. Unfortunately, run times for a single parameter set can take several minutes making it unsuitable for parameter inference where large numbers of repeated function evaluations are required. \cite{griffin_estimates_2014} later showed that a faster, reduced surrogate model can be used to infer these parameters. This surrogate made simplifying assumptions, such as the effects of seasonality and existing interventions being negligible. The observed data had to be carefully curated to minimise those effects. In contrast to a surrogate where a researcher arbitrarily chooses how to reduce model complexity for inference, we will introduce a deep learning solution.

Recent work has demonstrated the utility of learned surrogate models reducing the computational cost of simulation \cite{cameron_defining_2015,mckinley_approximate_2018,reiker_emulator-based_2021}. Data-driven methods could potentially learn more expressive, surrogates which allow researchers to relax their assumptions in parameter inference and use a wider range of data sets. \cite{cameron_defining_2015} proposed a functional regression method to map the functional space between model parameters and outputs using a library of pre-computed outputs. As the input space becomes large, more sophisticated methods have been employed to explore the sub-spaces most useful for the parameter inference task at hand, as reviewed by \cite{mckinley_approximate_2018}. \cite{reiker_emulator-based_2021} showed that a Bayesian optimisation workflow is effective in performing parameter inference on IBM outputs. They proposed training a surrogate and simultaneously performing inference, adaptively sampling parameter sets for training the surrogate which reduce the surrogate's predictive uncertainty in regions of the input space which are proximal to the best-fit parameters.

However none of the above approaches have developed surrogates designed for sequential inputs. Many epidemiological models use sequences as input, such as transmissibility \cite{rahimi_review_2021}, intervention usage \cite{griffin_estimates_2014} and contact matrices \cite{watson_global_2022}, and produce sequential outputs such as prevalence trajectories. This format makes them amenable to a sequence-to-sequence (seq2seq) modelling approach. Advances in seq2seq models have been very promising in machine translation \cite{vaswani_attention_2017} and time-series forecasting \cite{wen_transformers_2022}, but it is yet to be seen if they can be applied to the surrogate modelling of epidemiological models.

We hypothesised that a seq2seq based surrogate architecture could accurately predict IBM outputs for a combination of scalar and sequential parameters. Such a surrogate would make Bayesian inference feasible for long running observational data sets.

\subsection{Contributions}

We created a surrogate for a well known malaria model, allowing researchers to parameterise the surrogate for an arbitrary location and predict the malaria prevalence within milliseconds. Previous approaches have either restricted the input space to scalar parameters, or selected training samples for a specific inference task. As far as we know, no other infectious disease model surrogates are designed to take arbitrary sequence based parameters as input at prediction time.

We propose a formulation for the surrogate in section \ref{section:surrogate}. We outline a surrogate model structure and training method in sections \ref{section:structure} and \ref{section:sample} where we use synthetic parameter sets designed to explore the parameter space. We evaluated the surrogate on two benchmark parameter sets, in section \ref{section:evaluation}, to show that it generalises to longer parameter sequences and parameter sets generated from real world data. 

We then performed parameter inference using a Hamiltonian Monte Carlo algorithm on the learned surrogate. We estimated a parameter for a location in Senegal, from an 18 year sequence of intervention statistics. This is the first approach which would allow for inference using sequential parameters over an arbitrary time horizon and/or location without re-training the surrogate, which can reportedly take over a week per inference task, see \cite{reiker_emulator-based_2021}.

\section{Model of Malaria Transmission}
\label{section:model}

We first introduce a malaria model to provide context for the formulation of the surrogate model and to illustrate the complexity of an infectious disease model.

We used an implementation of an Individual-Based Model (IBM) of malaria originally published by \citeauthor{griffin_reducing_2010}. The IBM includes human, mosquito and Insecticide Treated Net (ITN) models. The human model simulates infectious bites, immunity, and the probability of detection by diagnostic tests. This model is used to produce malaria burden statistics and calculate the Force of Infection towards Mosquitoes (FOIM), as explained in section \ref{section:mosquito}. The mosquito model simulates the size of the infectious and susceptible mosquito populations given the FOIM and a time-series of rainfall. This model is used to calculate an Entomological Inoculation Rate (EIR), explained in section \ref{section:human}, which drives infections in the human model completing the cycle of infection between humans and mosquitoes. The bed net model modifies this cycle by reducing the number of bites in the protected human population and increasing mortality in the mosquito population. These effects are explained in section \ref{section:bednets}.

\subsection{Human Transmission}
\label{section:human}

\begin{figure}[ht]
  \includegraphics[scale=0.32]{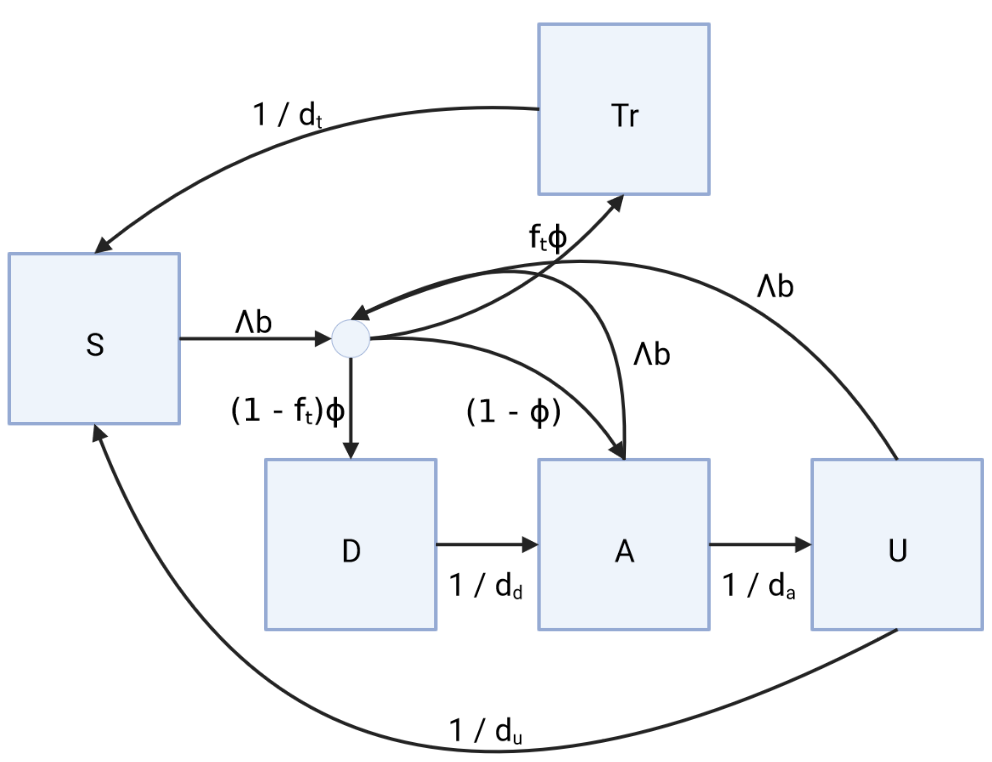}
  \caption{
      A human state transition diagram. Boxes represent human states and arcs transitions between the states, simulated with Bernoulli trials. $\Lambda$ represents the EIR, $f_t$ the probability of treatment, $b$ the probability of infection and $\phi$ the probability of clinical disease.
  }
  \label{fig:human}
  
\end{figure}

The human model, shown in figure \ref{fig:human}, simulates five infection states: susceptible (S), clinical disease (D), treated (T), asymptomatic infection (A), and subpatent infection (U). Subpatent infections are those not detected using routine malaria testing, instead requiring advanced diagnostic methods. At each time step, Bernoulli trials are simulated to transition individuals between these states at the parameterised rates. 

The Entomological Inoculation Rate (denoted $\Lambda$), is calculated as the number of infectious mosquito bites the population experiences over time. However, each individual experiences $\Lambda$ differently. Older individuals are bitten more often, due to their increased surface area. The age $a_i$ of each individual $i$ is sampled from an exponential distribution with mean $\mu_a$.

After an individual $i$ is bitten by an infectious mosquito, we simulate a possible infection. $b_i(t)$ represents the probability of blood-stage immunity failing, determining whether the individual will develop (at least) an asymptomatic infection. $\phi_i(t)$ represents likewise probabilities for clinical immunity and infection. For asymptomatic individuals, $q_i(t, a_i)$ represents the probability of detection by a routine diagnostic test, taking into account the individual's age. Each of these immunity functions are complex estimations based on each individual's previous exposures to bites and infection.

The diagnostic prevalence of malaria in the model $\mbox{PfPr}(t,l,u)$ is the proportion of individuals who would test positive in a routine diagnostic test, as defined in equation \ref{eq:prevalence}. This statistic is bounded between lower and upper age ranges $l$ and $u$, including all clinically diseased individuals and asymptomatic individuals who would be detected.

\begin{equation}
    \label{eq:prevalence}
    \begin{gathered}
        \mbox{PfPr}(t, l, u) = \frac{\sum_i \mbox{detected}(i, t) \times \mathds{1}_{l < a_i \leq u}}{\sum_i \mathds{1}_{l < a_i \leq u}} \\
        \mbox{detected}(i, t) =
        \begin{cases} 
          1 & \mbox{state(i, t)} = D \\
          q_i(t, a) & \mbox{state}(i, t) = A \\
          0  & \mbox{otherwise}
        \end{cases}
    \end{gathered}
\end{equation}

\subsection{Mosquito Modelling}
\label{section:mosquito}

Mosquitoes are modelled using a system of delay differential equations with early larval (E), late larval (L), pupal (P), susceptible (Sm), incubating (Em) and infectious (Im) compartments. Early (E) and Late (L) larval compartments represent the number of larvae soon after hatching. The size of these two compartments are dependent on a carrying capacity $K$ which can vary with rainfall $R(t)$, see equation \ref{eq:fourier}. The surviving larvae develop into Pupae (P) and half of the surviving pupae will develop into susceptible female mosquitoes.

\begin{gather}
    K^v = K^v_0 \frac{R(t)}{\bar{R}} \\
    R(t) = g_0 + \sum_{i=1}^3 g_i cos(2\pi t i) + h_i sin(2\pi t i) \label{eq:fourier}
\end{gather}

Susceptible female mosquitoes (Sm) are infected based on FOIM. FOIM is a driver of infection in the mosquito population, calculated by the number of bites each mosquito would have made on an infectious human. Infected mosquitoes will incubate in the (Em) state for a fixed period before themselves becoming infectious. If they survive, they will transition to the Infected (Im) state where they remain until death. The number of new E state larvae introduced each time step is proportional to the total number of adult female mosquitoes in the simulation.

The mosquito model is replicated for each species of mosquito in the simulation $v$. Users can parameterise the proportion of adult female mosquitoes in each species $\kappa^v$ by setting proportional values of the carrying capacity parameter, $K^v_0$. It is often convenient to parameterise $K^v_0$ using a baseline EIR parameter, $\Lambda_0$, or the expected EIR before interventions. \cite{griffin_is_2016} explains in detail how to calculate the carrying capacity from baseline EIR. 

\subsection{ITN Modelling}
\label{section:bednets}

Insecticide treated bednets (ITNs) are an effective malaria intervention which protect individuals while they are in bed. They are treated with an insecticide which repels or kills a mosquito when they attempt to feed on a human while in bed. Mosquitoes which are repelled will take longer to feed, indirectly affecting their mortality and reproduction rates. These responses, for various species and insecticides, have been examined in experimental hut trials, \cite{sherrard-smith_optimising_2022}. On the human side, EIR is reduced, since mosquitoes which have been repelled or killed before feeding no longer contribute towards infectious bites. We assume that bed nets are retained for an average of 5 years before being thrown away. The model can be parameterised with a sequence, $\nu(t)$, describing the proportion of the population who receive an ITN each year.

\subsection{Model Fitting Considerations}

Typically, malaria researchers would like to infer the unknown $\Lambda_0$ from observed prevalence data $\mbox{PfPr}(t,l,u)$, given known parameters about the transmission setting, such as $\mu_a,\kappa^v,\nu(t), g, h...$. However, simulating each individual, the mosquito population and ITN effects for several years can take several minutes, making parameter inference challenging. We also note that the individuals simulated by this model are synthetic and not measured directly by any real-world data source. Any fitting or inference process for this model will aggregate over individuals to predict population-level averages which can be compared to data.

In the next section we will formulate a model to predict these population-level aggregates directly from the IBM parameters, in order to eliminate a large amount of computation that is, for the purposes of fitting the model to data, unnecessary.

\section{Surrogate Model}
\label{section:surrogate}

Let $f:\mathcal X \to \mathcal Y$ be a model to be emulated. Assume that $\mathcal X \subset \mathbb R^k$ can be bounded to some set of plausible values. Our goal is to train a model $\tilde f_\theta : \mathcal X \to \mathcal Y$ such that $|\tilde f_\theta(\mathbf x) - f(\mathbf x)| < \epsilon$ for some suitably small $\epsilon >0$. We refer to $\tilde f_\theta$ as a surrogate model. Let $\mathcal L(\mathbf x, \mathbf y, \tilde f_\theta)$ be the loss of the surrogate model with respect to $\mathbf y$ (where $\mathbf y = f(\mathbf x)$).

The goal of surrogate modelling is to find a parameter set that replicates the original model accurately over the entirety of $\mathcal{X}$. That is, we want to find $\hat\theta$:

\begin{equation}
\hat \theta = \underset{\theta}{\mathrm{arg\;min}} \int_{\mathcal X}\mathcal L(\mathbf x, \mathbf y, \tilde f_\theta)p(\mathbf{x}) d\mathbf{x}.
\end{equation}

We have included a prior distribution over $\mathcal X$, but it is not strictly necessarily. Here, $p(\mathbf{x})$ serves as a measure of the relative importance of emulating $f(\mathbf{x})$ accurately. If $p$ is uniform, then we are ascribing equal importance to every possible parameter set.

\subsection{Input Domain}
\label{section:input}

We can translate the malaria IBM framing described in section \ref{section:model} by first defining $\mathcal X$.

To parameterise the surrogate for a particular location, we set $\kappa^v$, $\mu_a$, $g_i, h_i$ and $\nu(t)$ parameters to capture mosquito composition, demography, seasonality and ITN usage respectively. We defined a range for $\mu_a$ as 14.8 to 55.4 years based on the maximum and minimum average age statistics per country from World Population Review \cite{httpsworldpopulationreviewcom_2022_2022}. Ranges for Fourier coefficients $g_i, h_i$ are set between -10 and 10, to include the largest values found by fitting to rainfall data in sub-national administrative units in sub-Saharan Africa. Since $\kappa^v$ are relative proportions of three mosquito species, they are only valid in the 3-simplexes in $\mathbb R^3$.

A realistic range of $\nu(t)$ is between 0 and 0.8 for each year, as it becomes operationally difficult to distribute to ITNs to all individuals. We chose not to constrain the relationship between $\nu(t)$ and $\nu(t+\delta)$ as it could limit the the effectiveness of the surrogate in exploratory settings. However, parameterisations of $\nu(t)$ based on real world statistics are likely to have a gradual increase from 0 and plateau. In summary, $\nu(t)$ is defined as a bounded step function with $n \ge 0$ annually spaced knots.

One of the main unknowns for a each location is, $\Lambda_0$. An accurate estimate of this parameter would allow us to simulate a population of humans and mosquitoes which would recreate malaria transmission dynamics before any interventions were implemented. Previous estimates for countries in sub-Saharan Africa have ranged between 0.05 and 418 \cite{griffin_estimates_2014}, and model outputs suggest that higher values of $\Lambda_0$ have diminishing marginal effects on $\mbox{PfPr}$. Therefore we set a range for $\Lambda_0$ between 0 (exclusive) and 500.

To express $\mathcal X$ as a sequence, we formulated a sequence of $n$ vectors for each year, $t$, with time-invariant parameters ($\kappa^v, \mu_a, g_i, h_i, \Lambda_0)$ repeated at each time point, and the $\nu(t)$ selected for the corresponding time point. Each vector belongs to a subset of $\mathbb R^{13}$ bounded as described above.

\subsection{Output Range}

Let $(1, \ldots, T)$ be the sequence of years at which we wish to predict the prevalence of malaria. For any $\mathbf x \in \mathcal X$, let $f$ represent running the IBM with parameters $\mathbf x$ for $T$ years and evaluating $\mbox{PfPr}(d,l,u)$ for each day $d$ in each year $t$:

\begin{equation}
f(\mathbf x) = \left\{\left\{\mbox{PfPr}(d,u,l)\right\}_{d=1}^{365}\right\}_{t=1}^T
\end{equation}

Note that we have augmented the notation for the IBM run to include parameter sets. With this notation, $\mathcal Y$ is easy to define as the image of $f$:

\begin{equation}
\mathcal Y = \{f(\mathbf x):\mathbf x\in \mathcal X\},
\end{equation}

where $\mathcal X$ is defined as in the previous section. From inspecting equation \ref{eq:prevalence}, we get $\mathcal Y$ as a space of $T$-length sequences of vectors $\mathbb R^{365}$ bounded between 0 and 1.

\subsection{Surrogate Model Structure}
\label{section:structure}

With these definitions of $\mathcal{X}$ and $\mathcal{Y}$, the surrogate modelling task can be reduced to a sequence-to-sequence prediction problem. Sequence-to-sequence prediction is supported by a number of models, however we chose a recurrent neural network (RNN) for simplicity. The RNN should be able to exploit the relationships between time points to generalise for unseen parameter sets and longer time horizons. We used Mean Squared Error (MSE) as the loss function $\mathcal L$. We implemented a grid search select hyper-parameters which minimised MSE on the validation set, table \ref{table:hyper} lists the considered parameters.

\begin{table}[h]
\centering
      \begin{tabular}{|l | l |}
        \hline
        Hyper-parameter & Values \\ \hline
        Optimiser & \textbf{Adam}, RMSProp \\
        Optimiser Loss & MSE, \textbf{Log cosh}\\
        Batch size & 1000, \textbf{100}, 50\\
        Dropout & \textbf{0}, .1 \\
        RNN Layer & \textbf{LSTM}, GRU \\
        \hline
      \end{tabular}
\caption{Hyper-parameters used in the grid search. The selected parameter is emboldened}
\label{table:hyper}
\end{table}

Our RNN took each vector from the $\mathcal X$ sequence as input to its recurrent cells and output the corresponding vector from $\mathcal Y$. The parameters for $g_i$ and $h_i$ were expanded to full rainfall data $R(t)$ using equation \ref{eq:fourier}. The resulting inputs were standardised, removing the mean and scaling to unit variance. The final input layer had 383 recurrent cells, feeding 365 latent recurrent cells followed by a dense output layer of 365 cells.

\subsection{Sampling Training Data}
\label{section:sample}

Training a surrogate for the malaria IBM requires a large library of simulations covering a wide area of input space. We created training and validation data, with an 80/20 split for the surrogate model using a sequence length of $n = 16$. This would allow for the surrogate to learn the long term effects of ITNs on prevalence. We sampled 10,000 values from a 28 dimensional latin-hypercube, 16 for $\nu(t)$ and 12 for the remaining time-invariant parameters, and transformed the margins of each dimension to fit the bounds listed in section \ref{section:input}. The IBM was run at the sampled points, setting aside a warm-up period before activating the ITN parameters to allow the simulation to reach an equilibrium. We set a large population size for the IBM runs to minimise the stochasticity in $\mbox{PfPr}$ trajectories and treat them as deterministic outputs.

The simulation output was aggregated to calculate the prevalence of malaria among children aged 5 to 59 months for each year after a warm-up period. The age range of 5 to 59 months old was selected to align with statistics commonly found in observational data. 

\section{Surrogate Evaluation}
\label{section:evaluation}

\begin{figure}[ht]
  \includegraphics[width=.49\textwidth]{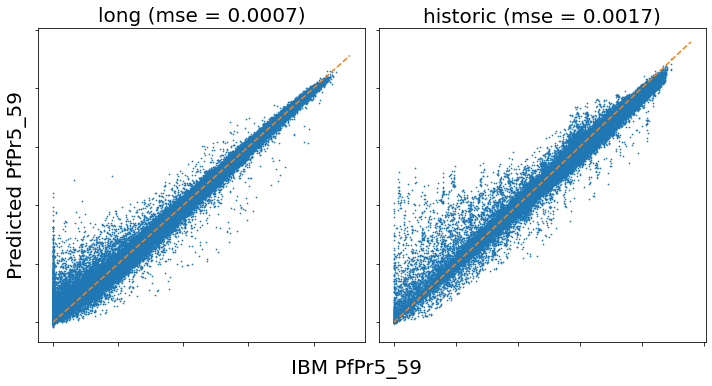}
  \caption{\protect{True versus predicted prevalence of malaria in children aged 5-59 months (\mbox{PfPr}5\_59) for the \textit{long} (left) and \textit{historic} (right) datasets where the true estimates come from the IBM model and the predicted estimates come from the surrogate. Each point represents the prevalence for one time step (a year) of one sample in the corresponding benchmark. Mean Squared Error (mse) was very low, as shown in the titles of the plots. }}
  \label{fig:pred}
\end{figure}

We evaluated the surrogate on two benchmark parameter sets. \textit{long} contains 5,000 samples generated similarly to the training set except for sequences with $n = 30$, testing the surrogate's ability to generalise to longer time horizons. \textit{historic} contains 5,000 samples where $\nu(t)$ was sampled from historic statistics for a randomly selected country in sub-Saharan Africa, published by \cite{bhatt_effect_2015}, testing the performance on parameters representative of global malaria modelling studies. After 100 epochs of the Adam optimiser, the MSE across the 2,000 out-of-sample simulations was $6.8 \times 10^{-3}$ on un-normalised prevalence values. The prediction error for these benchmarks is shown in figure \ref{fig:pred}.

We compared execution times for the original IBM, surrogate training, single prediction and batch prediction. The IBM executions were run on a cluster in batches of 10 on 1,000 dedicated cores with a 2GB memory reservation. The surrogate was trained on an Nvidia A1000 GPU, which took 229 $\pm$ 34 seconds (n = 3). Both single and batch predictions were made on an 8-core Intel CPU. While the IBM required just under 24 hours to run 10,000 samples, the batch surrogate required just over 3 seconds (approximately a 25,000-fold improvement). The timing statistics for an each run are listed in table \ref{table:timing}.

\begin{table}
   \begin{tabular}{|l|l|l|}
       \hline
       Task & Seconds per run & Std \\
       \hline
       IBM run & $1.76 \times 10^3$ & $7.69 \times 10^2$ \\
       Prediction (single) & $1.74 \times 10^{-3}$ & $2.84 \times 10^{-8}$ \\
       Prediction (batch) & $3.74 \times 10^{-4}$ & $6.12 \times 10^{-2}$ \\
       \hline
   \end{tabular}
  \caption{The timings per simulation for 10,000 simulations on the training and validation parameter sets}
  \label{table:timing}
\end{table}

\section{Parameter Inference}
\label{section:inference}

We used Tensorflow Probability \cite{httpswwwtensorfloworg_tensorflow_2022} to fit our model to observational data from the Demographic and Health Survey program \cite{httpsdhsprogramcom_dhs_2022}. The DHS responses included geo-location, age, and outcome of Rapid Diagnostic or Microscopy tests for respondents in Kolda, Senegal between 2008 an 2018. We included responses for respondents between 5 and 59 months old and weighted responses by population density data from World Bank data \cite{httpsdataworldbankorg_world_2022} to get a general picture of parasite prevalence over time in the region, $D_{\mbox{PfPr}5-59}$.

To estimate a value of $\Lambda_0$ which best explains $D_{\mbox{PfPr}5-59}$, we used a binomial likelihood function, see equation \ref{eq:likelihood}. We compared each participant's infection status $J^i$, an indicator variable, to the average predicted prevalence for the month in which they were observed, $\mbox{PfPr}(t^i)$. Our surrogate, parameterised with $\nu(t), g_i, h_i, \kappa^v$ values for Kolda, and aggregated at month $t^i$ provides quick estimates for $\mbox{PfPr}(t^i)$. We set a uniform prior for $\Lambda_0$ between 0 and 500.

\begin{equation}
\label{eq:likelihood}
L(\lambda^* | D_{\mbox{PfPr}5-59}) = \prod \mbox{PfPr}(t^i)^{J^i}(1 - \mbox{PfPr}(t^i))^{1 - J^i}
\end{equation}

We performed Markov Chain Monte Carlo using the HMC algorithm \cite{neal_mcmc_2011} with 10 parallel chains, 2,000 steps, 1,000 of which were discarded as warm-up, to sample from the resulting posterior. This inference took just over 11 minutes to execute on an 8-core Intel CPU.

\begin{figure}[ht]
  \includegraphics[scale=0.6]{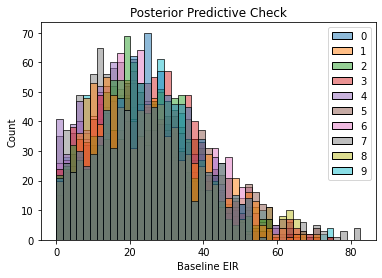}
  \caption{A posterior predictive check plotting $\Lambda_0$ samples from 10 MCMC chains. A previous rough estimate of baseline EIR in three Senegalese populations was 1, 20 and 200 (infectious bites per person per year) for the cities of Dakar, Ndiop and Dielmo \cite{trape_combating_1996}. This estimate of $23.2 \pm 13.7$  would place Kolda's baseline transmission alongside Ndiop.}
  \label{fig:ppc}
\end{figure}

\begin{figure}[ht]
  \includegraphics[scale=0.6]{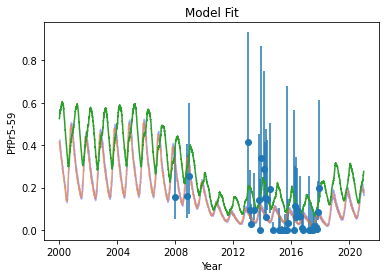}
  \caption{A plot of the final fit including the surrogate trajectory for the mean EIR (in orange). The IBM was then calibrated to the mean EIR using a root finding method and run to simulate the ITN distributions for Kolda, the prevalence for that run is plotted in green. Observations have been plotted in blue with their 95\% confidence interval, assuming a Binomial distribution. The observational data included samples from low intervention periods (2008) as well as high intervention periods (2013-2017) and was aggregated on a monthly time scale. }
  \label{fig:fit}
\end{figure}

The original IBM was then run with the mean estimate for comparison, see the posterior predictive check in figure \ref{fig:ppc} and resulting fit in figure \ref{fig:fit}. For this fit, there is a noticeable error between the surrogate and the original model. The seasonal cycle in prevalence is delayed and flattened, while the average prevalence is underestimated. However, the relative effects of ITNs appear similar between the models.

\section{Discussion}

We have demonstrated that it is feasible to perform inference on a highly detailed epidemic model with complex temporal dynamics. Seq2seq based surrogates can accurately predict trajectories for models with large parameter spaces and generalise transmission dynamics to very long time horizons. This opens up a wider range of data, including longitudinal, from less controlled studies for use in parameter inference in an epidemiological setting.

However, the inference in the previous section should not be used to inform real policy. Simplifying assumptions have been made which ignore the effects of non-ITN interventions and different diagnostic methods. Also, the accuracy of the surrogate varies for different types of model runs. Further work needs to be done to add detail to the surrogate model and ensure a high accuracy for the parameter space which is used for inference.

\subsection{A Surrogate for General Analysis}

The surrogate could be used as an exploratory tool, allowing researchers to begin modelling malaria transmission for arbitrary locations in seconds. Selective approaches, such as \cite{mckinley_approximate_2018,reiker_emulator-based_2021}, had designed their training set for a particular inference task, thus more training would be required to guarantee their published performance for each use case. In the case of \cite{reiker_emulator-based_2021}, training can take between 7 and 12 days.

\begin{figure}[ht]
  \includegraphics[scale=0.22]{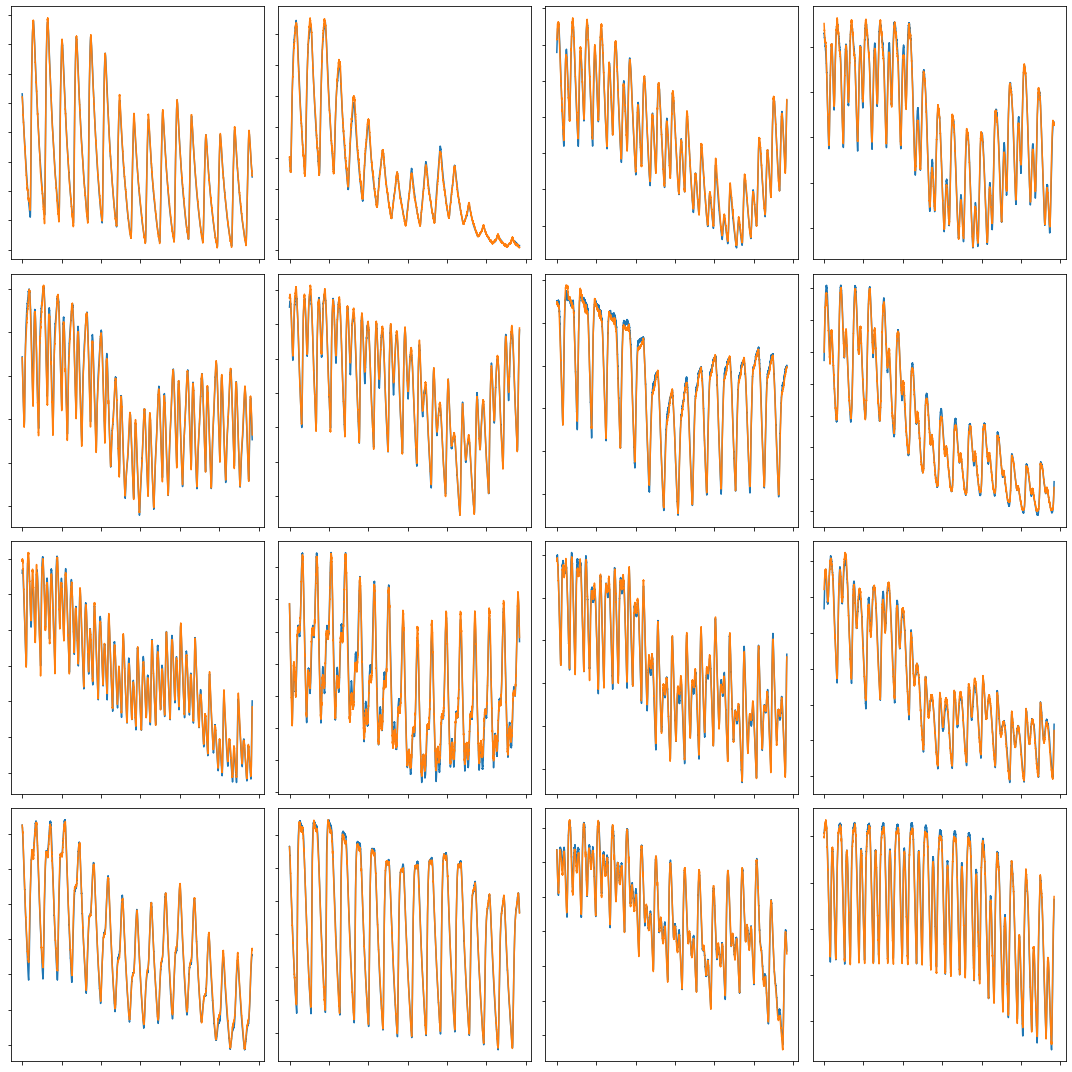}
  \caption{16 trajectories with the lowest MSE from the historic benchmark. The blue and orange trajectories show daily prevalence in simulation over 16 years for the IBM and surrogate models respectively. }
  \label{fig:best}
\end{figure}

The benchmarks showed that the surrogate could emulate the IBM for the vast majority of simulation scenarios. 99\% and 97\% of predictions had an MSE of less than $1 \times 10^{-2}$ for the \textit{long} and \textit{historic} benchmarks respectively. On inspection, we could see that short-term seasonal and long-term ITN transmission dynamics could be accurately re-created (see Figure \ref{fig:best}).

\begin{figure}[ht]
  \includegraphics[scale=0.22]{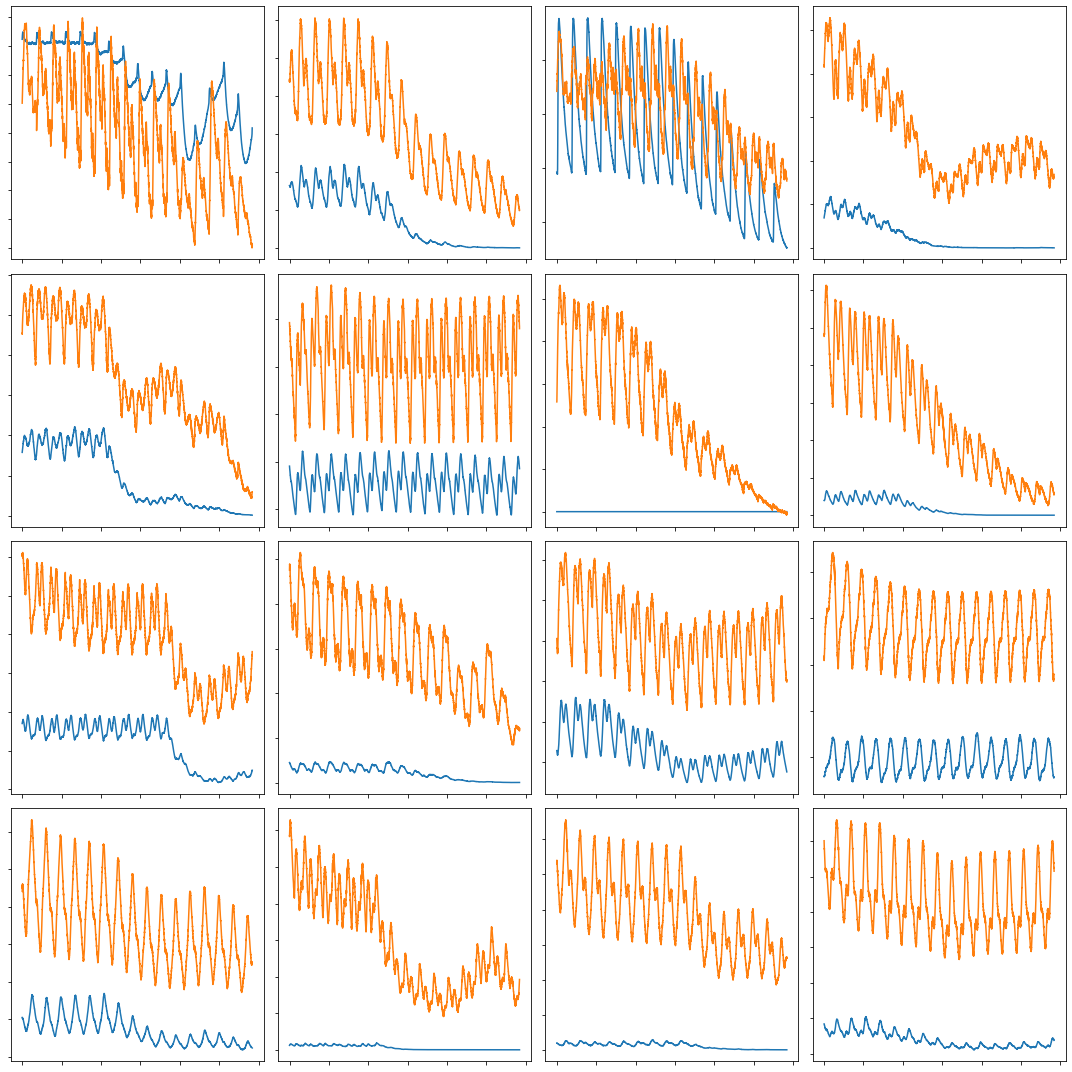}
  \caption{16 cases with the highest MSE from the historic benchmark. The blue and orange trajectories show daily prevalence in simulation over 16 years for the IBM and surrogate models respectively.}
  \label{fig:worst}
\end{figure}

Despite the benchmarks showing a small risk in surrogate prediction error, many predictions significantly deviated from the IBM outputs. Manual inspection reveals that the worst predictions for the historic benchmark were in scenarios with low or zero prevalence (see Figure \ref{fig:worst}). The performance of the surrogate would likely be improved by extending the training set with low transmission parameter sets, i.e. samples with smaller values of $\Lambda$.

A more general approach to managing the risk in surrogate prediction error could be uncertainty estimation. Unlike the approach in \cite{reiker_emulator-based_2021}, RNNs provide point estimates of the IBM outputs. Recent advances have shown that RNNs can be adapted to output probability densities, \cite{mackay_practical_1992,lakshminarayanan_simple_2017}. Such outputs could be used at prediction time to inform the researcher of the trust-worthiness of a prediction. They could also be used at training time to find samples which would reduce surrogate predictive uncertainty, \cite{bai_recent_2021}.

Finally, sharing a surrogate model is more energy efficient and equitable than an IBM. Low and middle income countries are disproportionately affected by infectious disease and are less able to finance high performance computing. Surrogate predictions require a small fraction of the IBM's computational requirements making analyses more feasible in the settings where they are most needed.

\subsection{Generality of the Workflow}

Despite the overall input domain being large, the vectors for each time point are low-dimensional. Over 50 malaria parameters were fixed to values sources from expert opinion and previous parameter inference studies \cite{sinka_global_2012,sherrard-smith_optimising_2022,griffin_estimates_2014}. If incorporated, it is unlikely that a latin-hypercube sampling strategy would produce the samples sufficient to learn the function $f$ over the larger input space. Developing a strategies to more intelligently sample the input space without sacrificing generality poses a challenging research problem.

The sequence structure is also limited in granularity. This surrogate has fixed time-step intervals to a year. However many infectious diseases, including malaria, are seasonal and so there is value in exploring the timing of interventions on a monthly or daily timescale. Reducing the time-step interval would increase the size of training sequences. For example, training this malaria surrogate with the daily time-step would increase the training sequences from $n = 16$ to $n = 5840$. More advanced seq2seq architectures could be considered here, such as those in \cite{cho_properties_2014,vaswani_attention_2017}, taking care to avoid the vanishing/exploding gradient problems.

\section{Conclusion}
We believe that the model presented here can be a valuable tool for guiding future malaria modelling, and serves as a proof of concept for the utility of surrogate modelling in infectious disease epidemiology. Given the limitations we have explored, future work will focus on improving the surrogate's fit and generalising the surrogate structure to adapt it to more varied prediction tasks.

\section*{Acknowledgements}

This research was funded in whole, or in part, by the Wellcome Trust [Grant number 220900/Z/20/Z]. For the purpose of open access, the author has applied a CC BY public copyright licence to any Author Accepted Manuscript version arising from this submission.

S.B. acknowledges support from the MRC Centre for Global Infectious Disease Analysis (MR/R015600/1), jointly funded by the UK Medical Research Council (MRC) and the UK Foreign, Commonwealth \& Development Office (FCDO), under the MRC/FCDO Concordat agreement, and also part of the EDCTP2 programme supported by the European Union.  SB is funded by the National Institute for Health Research (NIHR) Health Protection Research Unit in Modelling and Health Economics, a partnership between the UK Health Security Agency, Imperial College London and LSHTM (grant code NIHR200908). Disclaimer: “The views expressed are those of the author(s) and not necessarily those of the NIHR, UK Health Security Agency or the Department of Health and Social Care.” S.B. acknowledges support from the Novo Nordisk Foundation via The Novo Nordisk Young Investigator Award (NNF20OC0059309).  SB acknowledges support from the Danish National Research Foundation via a chair grant. S.B. acknowledges support from The Eric and Wendy Schmidt Fund For Strategic Innovation via the Schmidt Polymath Award (G-22-63345) which also supports GDC. SF acknowledges the EPSRC (EP/V002910/2).

\printbibliography

\end{document}


\maketitle

\section{Code}

The table below describes the various code attachments which were used to perform the analyses in this paper. Each package comes with installation and usage instructions at the root of their directories. \textit{msio} requires the malaria IBM model to be installed, made public by \cite{charles_malariasimulation_2022}.

\begin{tabular}{|l|p{.6\textwidth}|}
\hline
 Path & Description \\
\hline
 aiconf.sh & HPC script to create the training/validation set. \\
 aiconf\_test.sh & HPC script to create the \textit{long} benchmark. \\
 aiconf\_historic.sh & HPC script to create the \textit{historic} benchmark. \\
 miso & R package for sampling IBM runs, required by the "ai\_conf*.sh" scripts. \\
 fastms & Python package for training a surrogate based on msio outputs. \\
 observational.R & R script for extracting observational data from DHS. \\
 ITN.R & R script for extracting net usage data from the Malaria Atlas Project. \\
 IBM\_fit.R & R script for running the IBM at the inferred baseline EIR. \\
 ParameterInference.html & A html rendering of the jupyter notebook used to perform parameter inference. \\
 PredictiveError.html & A html rendering of the jupyter notebook used to analyse the surrogate's predictive error.\\
\hline
\end{tabular}

The surrogate was trained by executing the following fastms command, note that the seed was set to 42:

\begin{verbatim}
    python -m fastms.train [sample directory] 10000 .8 \
        [output directory] 100 42 --log=INFO
\end{verbatim}

The remaining random seeds (for the IBM sampling and parameter inference) are set by their respective HPC scripts and jupyter notebooks.

\section{Malaria model details}

\subsection{Individual EIR}

Individuals experience EIR differently based on their heterogeneity and age variables. Heterogeneity $\zeta_i$ is sampled from the log normal distribution, while age $a_i$ is sampled from the exponential distribution. The effective EIR experienced by each individual $\Lambda_i(a, t)$ at time $t$ is modelled relative to the EIR experienced by the complete human population $\Lambda(t)$.

\begin{equation}
    \label{eq:EIR}
    \begin{gathered}
        \Lambda_i(a, t) = \Lambda(t)\zeta_i(1 - \rho\exp(-\frac{a}{a_0})) \\
        log(\zeta_i) \sim N(\frac{-\sigma^2}{2}, \sigma^2) \\
        a \sim \exp{(\mu_a)}
    \end{gathered}
\end{equation}

\subsection{Immunity probabilities}

There are three functions representing immunity in the model. $b(t)$ and $\phi(t)$, which represent the probability of a blood-stage infection and progression to clinical disease, respectively, influence the infection transition rates, whereas $q_i(t, a)$ represents the probability of detection by a routine diagnostic test. These are calculated from acquired immunity variables $I_B, I_{CA}, I_D$, which are incremented on exposure to an infectious bite or on progression to clinical disease and later decay exponentially with rates $r_b$, $r_c$ and $r_d$. $\phi$ also takes into account a maternal immunity $I_{CM}$ which is set to a proportion of the individual's mother's clinical immunity at the time of birth. Maternal immunity also decays exponentially with rate $r_{cm}$. 

\begin{equation}
    \label{eq:immunity}
    \begin{gathered}
    b_i(t) = b_0 \left( b_1 + \frac{1 - b_1}{1 + \frac{I_B}{I_{B0}}^{k_b}} \right)\\
    \phi_i(t) = \phi_0 \left( \phi_1 + \frac{1 - \phi_1}{1 + \frac{I_{CA}(i, t) + I_{CM}(i, t)}{I_{C0}}^{k_c}} \right) \\
    q_i(t, a) = d_1 \left( d_1 + \frac{1 - d_{min}}{f_D(i, a)\frac{1 + I_D(i, t)}{I_{D0}}^{k_D}} \right)
    \end{gathered}
\end{equation}

\subsection{Delay differential equation model of mosquitoes}

The mosquito population is modelled using the differential equations listed in  \ref{eq:vectors}. They include a delay of $\tau$ to model an incubation period before mosquitoes become infected.

\begin{equation}
    \label{eq:vectors}
    \begin{gathered}
        \frac{dE^v}{dt} = \beta_{egg}^v(t) M - \frac{E^v}{de} - \mu_e E^v \left(1 + \frac{E^v + L^v}{K^v}\right) \\
        \frac{dL^v}{dt} = \frac{E^v}{de} - \frac{L^v}{dl} - \mu_l L^v \left(1 + \gamma \frac{E^v + L^v}{K^v}\right)  \\
        \frac{dP^v}{dt} = \frac{L^v}{dl} - \frac{P^v}{dp} - \mu_p P^v \\
        \frac{dS^v_m}{dt} = -\Lambda_M^v S_m + \frac{P^v}{2 dp} - \mu^v S^v_m \\
        \frac{dE^v_m}{dt} = \Lambda_M^v S^v_m +
            - \Lambda_M^v(t - \tau) S^v_m(t - \tau) \exp(-\mu^v\tau) - \mu^v E^v_m \\
        \frac{dI^v_m}{dt} = \Lambda_M^v(t - \tau) S^v_m(t - \tau) \exp(-\mu^v\tau) - \mu^v I^v_m
    \end{gathered}
\end{equation}

\subsection{Calculation of FOIM}

Susceptible female mosquitoes (Sm) are infected based on FOIM, or $\Lambda_M$, expressed in equation \ref{eq:FOIM}, which is in turn related to the species specific biting rate $\alpha^v$ and the infectiousness of the human population $\sum_i c(i, t)$. Infected mosquitoes will experience a delay $\tau$ in the Em state before becoming infectious. If they survive, they will transition to the Infected (Im) state where they remain until death. The number of new E state larvae introduced each time step, $\beta_{egg}$ is proportional to the total number of adult female mosquitoes in the simulation.

\begin{equation}
\label{eq:FOIM}
\Lambda^v_M(t) = \alpha^v \sum_i c(i, t) \pi(i)
\end{equation}

where,

\[
c(i, t) =
\begin{cases} 
  0  & state(i, t) \in \{S, T\} \\
  c_D & state(i, t) = D \\
  c_U & state(i, t) = U \\
  c_A(i, t) & state(i, t) = A
\end{cases}
\]

where,

\[  c_A(i, t) = c_U + (c_D - c_U) q_i \gamma^1 \]

\subsection{ITN model}

We model the probability of a feeding without death and repulsion, as $w_i^v, z_i^v$, in terms of the properties of the bed nets and mosquito behaviours, see equations \ref{eq:nets}. The probability of survival and repulsion for a particular bed net, or $s_N, r_N$, have been estimated from experimental hut trial data by \cite{sherrard-smith_optimising_2022}. These probabilities are reduced over time as the insecticide wears off through use. The probability that a mosquito of species $v$ feeds on humans in bed, $\Phi_B^v$ is estimated for common infectious species of mosquito in \cite{sherrard-smith_optimising_2022}. High bed net usage, the proportion of individuals protected by bed nets, reduces the successful feeding attempts on the population, increases mosquito mortality, $\mu^v$, and the rate at which an individual is bitten. The model can be parameterised with a time-series, $\nu(t)$, describing the proportion of the population who receive an ITN each year.

\begin{equation}
    \label{eq:nets}
    \begin{gathered}
        w^v_i = 1 - \Phi_B^v + \Phi_B^vs_N\\
        z^v_i = \Phi_B^vr_N
    \end{gathered}
\end{equation}

\subsection{Model parameters}

IBM most model parameters were taken as the median values of the Bayesian analysis done by \cite{griffin_is_2016}, listed in the table below:

\begin{longtable}{|l |p{.6\textwidth}| c |} 
 \hline
\label{table:parameters}
 Parameter & Description & Value \\ [0.5ex] 
 \hline
 \multicolumn{2}{|l}{Fixed state transitions:} & \\
 \hline
    dd & The delay for humans to move from state D to A & 5 \\

    dt & The delay for humans to move from state T to S & 5 \\

    da & The delay for humans to move from state A to U & 200 \\

    du & The delay for humans to move from state U to S &  110 \\

    del & The delay for mosquitoes to move from state E to L &  6.64 \\

    dl & The delay for mosquitoes to move from state L to P &  3.72 \\

    dpl & The delay mosquitoes to move from state P to Sm &  0.643 \\

    mup & The rate at which pupal mosquitoes die &  0.249 \\

    mum & The rate at which developed mosquitoes die &  0.1253333 \\

\hline
 \multicolumn{2}{|l}{Immunity decay rates:} & \\
\hline

    rm & Decay rate for maternal immunity to clinical disease &  67.6952 \\

    rvm & Decay rate for maternal immunity to severe disease &  76.8365 \\

    rb & Decay rate for acquired pre-erythrocytic immunity &  3650 \\

    rc & Decay rate for acquired immunity to clinical disease &  10950 \\

    rva & Decay rate for acquired immunity to severe disease &  10950 \\

    rid & Decay rate for acquired immunity to detectability &  3650 \\

\hline
 \multicolumn{2}{|l}{Pre-erythrocytic infection:} & \\
\hline

    b0 & Maximum probability due to no immunity &  0.59 \\

    b1 & Maximum reduction due to immunity &  0.5 \\

    ib0 & Scale parameter &  43.9 \\

    kb & Shape parameter &  2.16 \\

\hline
 \multicolumn{2}{|l}{Probability of clinical infection:} & \\
\hline

    phi0 & Maximum probability due to no immunity &  0.792 \\

    phi1 & Maximum reduction due to immunity &  0.00074 \\

    ic0 & Scale parameter &  18.02366 \\

    kc & Shape parameter &  2.36949 \\

\hline
 \multicolumn{2}{|l}{Probability of detection:} & \\
\hline

    fd0 & Time-scale at which immunity changes with age &  0.007055 \\

    ad & Scale parameter relating age to immunity &  7993.5 \\

    gammad & Shape parameter relating age to immunity &  4.8183 \\

    d1 & Minimum probability due to immunity &  0.160527 \\

    id0 & Scale parameter &  1.577533 \\

    kd & Shape parameter &  0.476614 \\

\hline
 \multicolumn{2}{|l}{Immunity boost grace periods:} & \\
\hline

    ub & Period in which blood immunity cannot be boosted &  7.2 \\

    uc & Period in which clinical immunity cannot be boosted &  6.06 \\

    uv & Period in which severe immunity cannot be boosted &  11.4321 \\

    ud & Period in which detection immunity cannot be boosted &  9.44512 \\

\hline
 \multicolumn{2}{|l}{Infectivity towards mosquitoes:} & \\
\hline

    cd & Infectivity of clinical disease towards mosquitoes &  0.068 \\

    $\gamma^1$ & Parameter for infectivity of asymptomatic humans &  1.82425 \\

    cu & Infectivity of sub-patent infection &  0.0062 \\

    ct & Infectivity of treated infection &  0.021896 \\

\hline
 \multicolumn{2}{|l}{Unique biting rate:} & \\
\hline

    a0 & Age dependent biting parameter &  2920 \\

    rho & Age dependent biting parameter &  0.85 \\

    $\sigma^2$ & Heterogeneity parameter &  1.67 \\

\hline
 \multicolumn{2}{|l}{Mortality parameters:} & \\
\hline

    pcm & New-born clinical immunity relative to mother's &  0.774368 \\

    me & Early stage larval mortality rate &  0.0338 \\

    ml & Late stage larval mortality rate &  0.0348 \\

\hline
 \multicolumn{2}{|l}{Carrying capacity parameters:} & \\
\hline

    $\gamma$ & effect of density dep. on late relative to early instars &  13.25 \\

\hline
 \multicolumn{2}{|l}{Incubation periods:} & \\
\hline

    de & Duration of the human latent period of infection &  12 \\

    delay\_gam & Lag from parasites to infectious gametocytes &  12.5 \\

    dem & incubation period in mosquito population model &  10 \\

\hline
 \multicolumn{2}{|l}{Vector biology:} & \\
\hline

    beta & The average number of eggs laid per mosquito per day &  21.2 \\

    foraging\_time & Time spent taking blood meals &  0.69 \\

    \hline
\end{longtable}

For Kolda, we used the seasonality, vector composition and demography parameters below:

\begin{tabular}{|l|l|}
\hline
 Parameter & Value\\
\hline
 $g_0$ & 2.574537504 \\
 $g_1$ & 3.489987862 \\
 $g_2$ & 0.657536368 \\
 $g_3$ & 0.565366933 \\
 $h_1$ & -2.39714463 \\
 $h_2$ & 2.189027167 \\
 $h_3$ & -0.565822876 \\
 $\kappa_1$ & .25 \\
 $\kappa_2$ & .25 \\
 $\kappa_3$ & .5 \\
 $\mu_a$ & 18.5 \\
\hline
\end{tabular}

 ITN usage, $\nu(t)$, was set using the values below:

\begin{tabular}{|l|l|}
\hline
 Year &      ITN usage \\
\hline
 2000 & 0.046681 \\
 2001 & 0.034206 \\
 2002 & 0.037521 \\
 2003 & 0.033482 \\
 2004 & 0.020535 \\
 2005 & 0.030971 \\
 2006 & 0.122021 \\
 2007 & 0.168003 \\
 2008 & 0.214104 \\
 2009 & 0.426100 \\
 2010 & 0.594891 \\
 2011 & 0.653355 \\
 2012 & 0.480221 \\
 2013 & 0.525787 \\
 2014 & 0.588837 \\
 2015 & 0.645039 \\
 2016 & 0.788848 \\
 2017 & 0.715216 \\
 2018 & 0.349282 \\
 2019 & 0.515891 \\
 2020 & 0.455649 \\
\hline
\end{tabular}

\section{Observational data}

Aggregated test results for DHS respondents in Kolda between 5 and 59 months old. The number of respondents are weighted by population density, giving non-integer results. The diagnostic prevalence is calculated as: $\frac {Positive}{Positive + Negative}$

\begin{tabular}{|l|l|r|r|r|}
\hline
Year &  Month &    Negative &   Positive &    Prevalence \\
\hline
 2008 &      1 & 107.425282 & 20.024040 & 0.157114 \\
 2008 &     11 &  89.312360 & 17.140296 & 0.161013 \\
 2008 &     12 &  82.889441 & 28.390398 & 0.255126 \\
 2013 &      1 &  52.896958 & 37.128040 & 0.412419 \\
 2013 &      3 &  59.598213 &  6.221322 & 0.094521 \\
 2013 &      4 &  35.675691 &  0.946405 & 0.025842 \\
 2013 &      6 &  71.836294 &  7.404132 & 0.093439 \\
 2013 &     10 &  21.607676 &  3.612997 & 0.143255 \\
 2013 &     11 &  13.498408 &  0.000000 & 0.000000 \\
 2013 &     12 &  18.322859 &  9.466030 & 0.340641 \\
 2014 &      2 &  21.188314 &  8.574896 & 0.288104 \\
 2014 &      3 &  43.127789 &  7.645989 & 0.150589 \\
 2014 &      4 &   6.395200 &  0.399700 & 0.058824 \\
 2014 &      5 &  28.688672 &  4.848680 & 0.144576 \\
 2014 &      7 &  50.286429 & 12.076372 & 0.193647 \\
 2014 &     10 &  95.122290 &  0.000000 & 0.000000 \\
 2015 &      3 &  33.793043 &  0.000000 & 0.000000 \\
 2015 &      4 &  53.376397 &  0.000000 & 0.000000 \\
 2015 &      5 & 182.046854 &  0.000000 & 0.000000 \\
 2015 &      8 &  46.592927 &  0.000000 & 0.000000 \\
 2015 &      9 &   3.274010 &  0.000000 & 0.000000 \\
 2015 &     10 &  29.622070 &  0.961008 & 0.031423 \\
 2016 &      3 &   4.433990 &  0.000000 & 0.000000 \\
 2016 &      4 &  76.578173 &  9.702714 & 0.112455 \\
 2016 &      5 &  10.479957 &  0.705492 & 0.063072 \\
 2016 &      6 &  44.745397 &  4.729929 & 0.095602 \\
 2016 &      8 &  41.429804 &  3.011445 & 0.067762 \\
 2016 &     10 &  56.193909 &  0.621489 & 0.010939 \\
 2016 &     11 &  33.507003 &  0.000000 & 0.000000 \\
 2017 &      4 &  12.589093 &  0.000000 & 0.000000 \\
 2017 &      5 &  25.789797 &  0.000000 & 0.000000 \\
 2017 &      7 &  40.005915 &  0.000000 & 0.000000 \\
 2017 &      8 & 108.341670 &  2.214433 & 0.020030 \\
 2017 &      9 & 106.978260 &  1.423310 & 0.013130 \\
 2017 &     10 & 127.933200 &  0.740660 & 0.005756 \\
 2017 &     11 &  86.161314 &  7.982419 & 0.084790 \\
 2017 &     12 &  11.653305 &  2.861636 & 0.197151 \\
\hline
\end{tabular}

\bibliography{aiconf.bib}